\documentclass[sigconf]{acmart}

\AtBeginDocument{%
  \providecommand\BibTeX{{%
    \normalfont B\kern-0.5em{\scshape i\kern-0.25em b}\kern-0.8em\TeX}}}

\setcopyright{none}
\copyrightyear{}
\acmYear{}
\acmDOI{}
\acmConference[DSHealth '20]{DSHealth '20: 2020 KDD Workshop on Applied Data Science for Healthcare}{August 24, 2020}{San Diego, CA}
\acmBooktitle{}
\acmPrice{}
\acmISBN{}

\usepackage[caption=false]{subfig}
\graphicspath{{images/}}

\makeatletter
\def\@copyrightspace{\relax}
\makeatother
\begin{document}

\title{A Canonical Architecture For Predictive Analytics on Longitudinal Patient Records }


\author{Parthasarathy Suryanarayanan$^{1*}$, Bhavani Iyer$^1$, Prithwish Chakraborty$^1$, Bibo Hao$^1$, 
       Italo Buleje$^1$, Piyush Madan$^1$, James Codella$^1$, Antonio Foncubierta$^1$, Divya Pathak$^1$, Sarah Miller$^1$,\and 
       Amol Rajmane$^2$, Shannon Harrer$^2$, Gigi Yuan-Reed$^2$, Daby Sow$^1$} 
    \affiliation{ 
      \institution{$^1$IBM Research; $^2$IBM Watson Health}
      \state{Corresponding Author: \url{psuryan@us.ibm.com}}
    }

\renewcommand{\shortauthors}{P Suryanarayanan, et al.}
\settopmatter{printacmref=false, printccs=true, printfolios=true}

\begin{abstract}
Many institutions within the healthcare ecosystem are making significant investments in AI technologies to optimize their business operations at lower cost with improved patient outcomes. Despite the hype with AI, the full realization of this potential is seriously hindered by several systemic problems, including data privacy, security, bias, fairness, and explainability. In this paper, we propose a novel canonical architecture for the development of AI models in healthcare that addresses these challenges. This system enables the creation and management of AI predictive models throughout all the phases of their life cycle, including data ingestion, model building, and model promotion in production environments. This paper describes this architecture in detail, along with a qualitative evaluation of our experience of using it on real world problems.  
\end{abstract}




\maketitle

\section{Introduction}\label{sec:introduction}
Recent surveys run by many organizations, research advisory companies, government entities and media outlets have been pointing at the importance and potential of AI technologies to transform the way healthcare is delivered. In~\cite{hit_AI_adoption}, authors noted that 91\% of healthcare stakeholders believe that adoption of AI will lead to improved patient access to care. Despite such an optimistic outlook about the impact of AI in healthcare, many fear that several barriers need to be overcome to fulfill this potential. These barriers are driven by several factors, including the need for more standardized and interoperable ways to access, manage and maintain data and AI models, the need to provide trust in AI modeling through complete transparency, explainability, data and model provenance, and the need for enhanced security and privacy around the secondary reuse of patient data~\cite{hit_whouse}. Even though partial solutions for addressing each of these individual issues are available, a system that brings all of them together into a cohesive architecture for healthcare is novel and is our main contribution.        

To this end, we propose a canonical architecture for the complete management of predictive healthcare AI applications throughout all phases of their life cycle, such as data ingestion, model building, and model promotion into production environments. The architecture is designed to accommodate trust and reproducibility as an inherent part of the AI life cycle and support the needs for a deployed AI system in healthcare. In what follows, we start with a crisp articulation of challenges that we have identified to derive the requirements for this architecture. We then follow with a description of this architecture before providing qualitative evidence of its capabilities in real world settings. 

\section{Challenges}\label{sec:challenges}
While AI offers powerful tools for building useful complex prediction systems quickly, it is common to incur massive ongoing maintenance costs in real-world AI systems. Many of the production AI systems are inherently brittle due to various reasons from underutilized data dependencies to lack of code-reuse between training and inference pipelines~\citep{sculley2015hidden}. Systems such as Facebook’s FBLearner \citep{dunn2016introducing}, Uber’s Michelangelo \citep{hermann2017meet} and DataBricks MLFlow \citep{zaharia2018accelerating} have developed approaches and platforms to manage machine learning workflows for general use cases. However, healthcare workflows pose additional challenges while incorporating AI. They necessitate new approaches for each step from data collection, model development to validation, deployment and monitoring~\citep{liuGoogleAIBlog,piaynykhNatMachIntell2020}. Some of the domain specific challenges are given below. 
\begin{enumerate}
    \item Integration of data from multiple sources such as insurance claims, clinical data from EHRs, provider profiles, population statistics, social and community data, oncologies and other curated resources is essential for generating useful machine learning features. 
    \item Inference processes are often complex, involving multiple models to support explainable, actionable, and bias mitigated predictions. 
    \item Computing prediction at low-latency in the context of long-term historical event data is important.
    \item High degree of accuracy is required of predictions for critical decision-making which in turn requires continuous monitoring and tuning of model performance \citep{liuGoogleAIBlog}.
    \item The system should provide generated results in a transparent manner to drive trust; it should also be able to provide explanations to end users on how these results were obtained.
    \item Adherence to security and privacy regulations such as HIPAA and protecting the AI modeling from various attacks.
\end{enumerate}
While these challenges are present in other domains, addressing their aggregation is imperative to stand up production AI systems in healthcare. 

\begin{figure*}
    \subfloat[Component diagram\label{fig:logical_view_component_diagram}]{%
      \includegraphics[height=8cm,width=.49\linewidth]{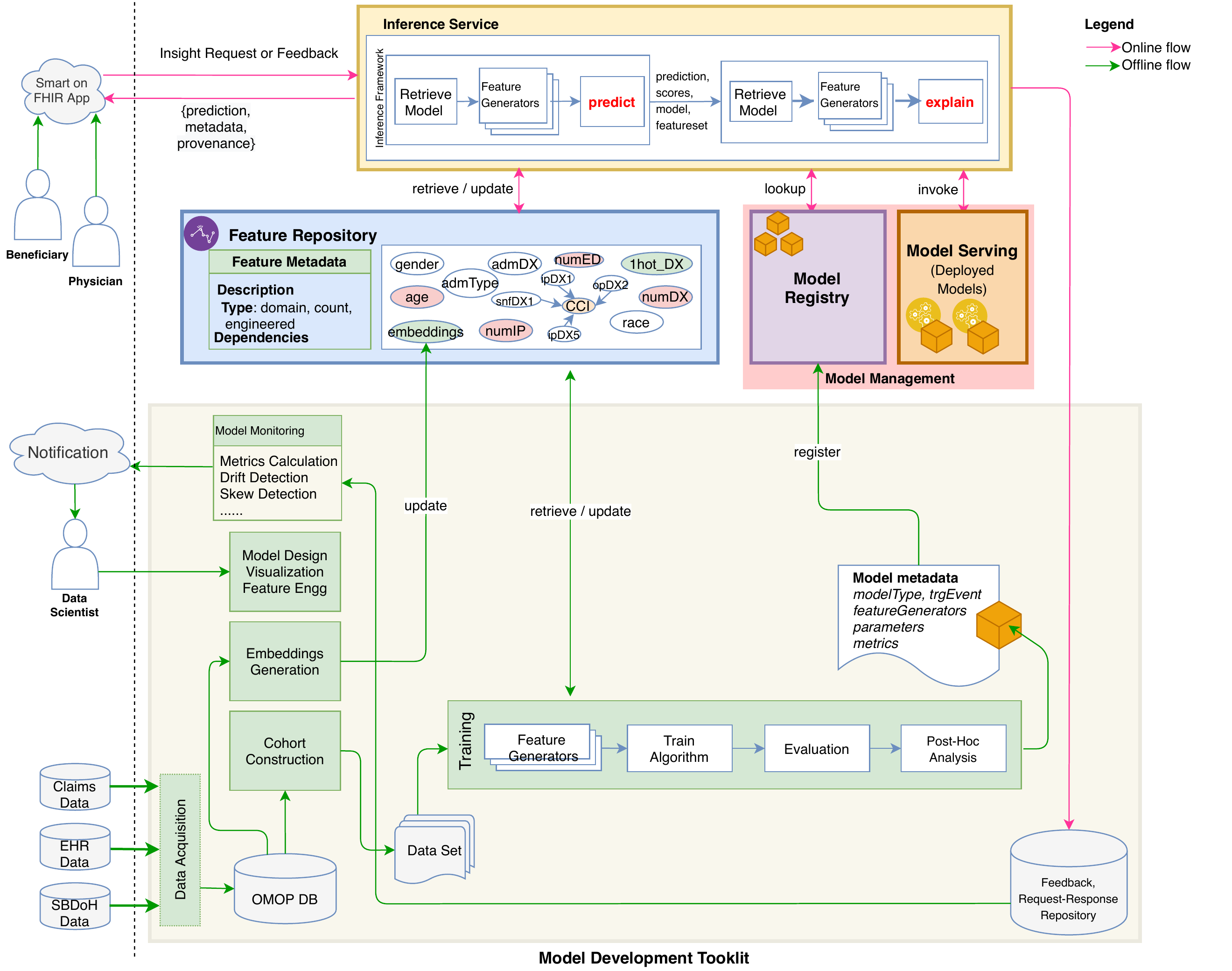}%
    }\hfill
    \subfloat[AI life-cycle management\label{fig:ai-life-cycle}]{%
      \includegraphics[height=8cm,width=.49\linewidth]{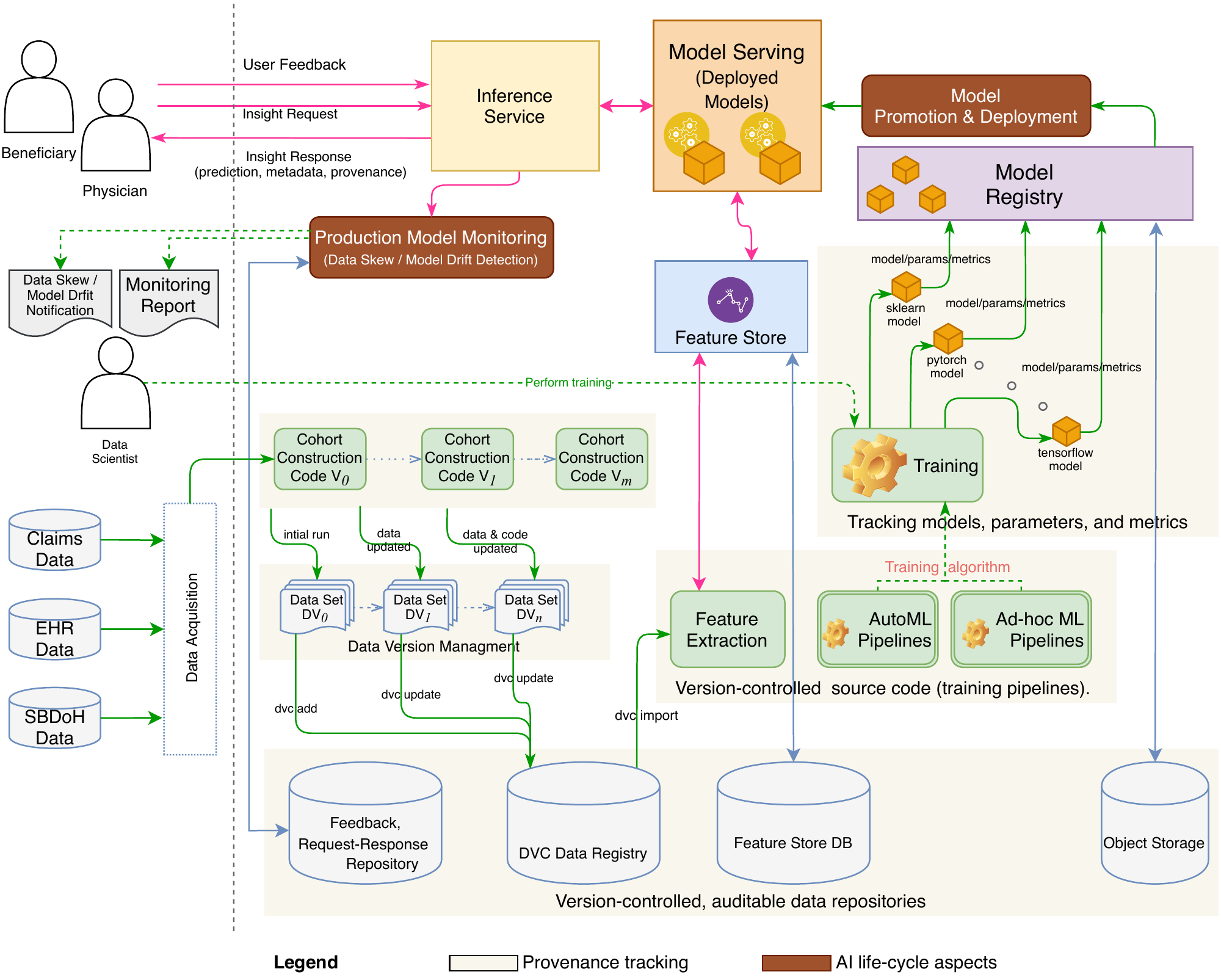}%
   }
    \caption{A Canonical Architecture for Predictive Analytics on EHR. (a) The logical view of the four main components of the architecture: (top$\rightarrow$bottom) \textit{Inference Frame-work}, \textit{Model Management Subsystem}, \textit{Feature Repository}, and \textit{Model Development Toolkit}. (b) The process view depicting the various stages of the AI life-cycle enabled by the architecture.}
    \label{fig:ref-arch}
\end{figure*}
\section{Desiderata for Architecture}\label{sec:requirements}
 To meet the above challenges we identify the following as desirable characteristics of a solution architecture. 

\textbf{Modularity}: The system should have the ability to incorporate new features through extension rather than modification. Building an AI solution involves enabling collaboration between cross-functional teams with diverse set of skills needed to handle data, code and model development. To enable such collaborative research and development, the architecture should be highly modular with a well-defined AI life-cycle management process.

\textbf{Trust \& Transparency}: Some desired properties are:
\begin{enumerate}
\item  The reliability of predictive behavior of models in real world deployments varies when confronted with real world data. Thus, there is a need for continuous monitoring of model performance~\cite{breck2016s} and for a systematic approach to address model staleness: continuous model evaluation, "training/inference skew" detection, and model drift to trigger retraining.
\item Maintaining provenance of data, models, and software is crucial for reproducibility and monitoring model performance in production.
\item The system should have a mechanism to reliably consume and incorporate user feedback to improve models.
\item The system should support state-of-the-art explainable AI modeling technologies.
\end{enumerate}

\textbf{Ease of Integration}: The system should integrate seamlessly with existing clinician workflows within existing tools and applications that are already familiar to clinicians\citep{liuGoogleAIBlog}. Also, the system should support integration of diverse data sources.

\textbf{Healthcare Interoperability}: Support for open standards for healthcare such as FHIR.


\textbf{Security \& Privacy}: The system must adhere to security and privacy regulations such as HIPAA.

\textbf{Performance}: The architecture should ensure that non-functional requirements for performance such as throughput, latency, or memory usage are met 
in addition to the requirements above
\citep{qayyum12020RobustML}.


In the next section, we describe a system that builds upon the ideas from ~\citep{cd4ml} to meet the architectural goals mentioned above.

\section{Architecture}\label{sec:architecture}
Our proposed architecture consists of four subsystems: 1. Inference Framework, 2. Model Management Subsystem, 3. Feature Repository and 
4. Model Development Toolkit.

\subsection{Inference Framework}\label{sec:prediction-framework}
The \textit{Inference Framework}, is responsible for the generation of insights for end user consumption. It runs inside the \textit{Inference Service}, and organizes the various software components into a cohesive set of modules through contracts. These modules are best described through their interactions with each other during the request processing control flow.
\begin{enumerate}
    \item Upon receiving request for insight through the \textit{API Gateway}, the framework calls the \textit{Model Registry} to fetch the model specification (\textit{ModelSpec}). 
    \item The returned \textit{ModelSpec} consists of
        (a) a handle to the micro-service corresponding to the deployed model in the \textit{Model Serving} runtime environment,
        (b) a list of machine learning feature generation components (\textit{Feature Generators}) that were used during training,
        (c) a reference to the list \textit{Model Metadata} components that can generate metadata associated with the prediction such as explainability, 
        actionability and robustness,
        (d) provenance information that details the model algorithm, training inputs, parameters and metrics.
    \item The framework then executes a sequence of steps based on the \textit{ModelSpec} and the incoming request parameters to generate a prediction and prediction metadata.
    \item A response is composed and returned to the client. The inference framework also logs the request-response pair in a repository for later use in model re-evaluation.  
\end{enumerate}    
The system allows for capturing user feedback. On receipt of user feedback for a previous prediction through the \textit{API Gateway}, the feedback along with all metadata associated with feedback (such as the state within the clinical workflow when the feedback was submitted) is logged in to a \textit{Feedback Repository}.

\subsection{Model Management Subsystem}\label{sec:model-management-subsystem}
The Model Management Subsystem manages many elements of the model life-cycle. It is used to register models after training, retrieve model specification and provenance, and to execute models at runtime. It is built on top of MLflow~\citep{zaharia2018accelerating} and consists of following components:
(1) \textit{Model Registry} where the model, its specifications, metrics and provenance are registered. ``Training pipeline'' uses the \textit{Model Registry} to log the model specifications, feature generators, prediction metadata generators, metrics, and provenance and register the model. It is also used by the  \textit{Inference Framework} to retrieve the best model for the machine learning task. 
(2) \textit{Model Serving} allows models to be turned into micro-services and inference is run within the service usually via a request-response paradigm.

\subsection{Feature Repository}\label{sec:feature-repository-subsystem}

Feature generation, the process of transforming raw input data into features in formats expected by the machine learning algorithm is needed both during training and real-time inference. In most large scale machine learning projects, feature generation is done by a diverse team that utilizes a variety of methods, tools, and implementation approaches.  How features are generated, maintained, and made available has an impact on the complexity of the system. Improperly managed feature generation can affect feature discovery,  reuse, and overall the reliability of predictions \citep{miao2018provdb} resulting in technical debt~\citep{sculley2015hidden}. The concept of a \textit{Feature Repository} was introduced by Uber \citep{hermann2017meet} and since gained a prominence. Implementations are typically based on a NoSQL database  and service APIs to retrieve, add update feature data \citep{condenast_feature_stores_2019}.
The \textit{Feature Repository} provides several benefits as follows:

\textbf{Provides provenance}: The \textit{Feature Repository} supports storage and retrieval of historical, versioned feature values. 

\textbf{Aids modularity}: It enables standardization of definition, storage and access to feature data, promoting reuse and less duplication.

\textbf{Accelerates innovation}: Easy discovery of feature sets can jump start machine learning models, increase learning efficiency and lowering model development costs. This component provides interfaces and visualization tools to for data  exploration, error analysis, and model tuning by data scientists.

\textbf{Improves run-time performance}: It reduces latency of prediction by using pre-computed features that can be queried against a database  instead of through a complex operation involving aggregations of large amounts of historical data. 
This also improves the throughput of data ingestion by capturing and exploiting the inter-dependencies between features to trigger re-computation or incremental updates. \textit{Feature Generators} can also be grouped together for efficient execution.  
\subsection{Model Development Toolkit}\label{sec:model-development-toolkit}
Many teams build custom tools and use adhoc ways to address requirements in the AI life-cycle resulting in massive 
technical debt \cite{sculley2015hidden}. To address this, our solution incorporates a comprehensive toolkit for model development with 
integrated AI life-cycle management. It accelerates development and reduces maintenance efforts by promoting repeatable workflows. It promotes reuse while providing the flexibility needed for data scientists to innovate.  The model development process has several stages:

\textbf{Data Acquisition}: This step provides an interface for ingesting data from a variety of sources, 
    such as EMR, claims, and Social and Behavioral Determinants of Health (SBDoH), for periodic retraining of the models as necessary. 
    This also transforms and normalizes the data into a common data model for downstream 
    tasks.

\textbf{Cohort Construction}: The first step in model building is to construct a cohort. Cohort construction interfaces 
    with the raw data and the \textit{Feature Repository}.  Domain features that can be directly extracted from the data such as 
    demographics, admission diagnosis, can be added to the \textit{Feature Repository}. In a production setting, 
    cohorts may be extended in two ways: (1) addition of new types of data, (e.g., new types of claims, or claims from different time periods), (2) defining new target events (e.g. unplanned admission, opiate use disorder, maternal morbidity, etc.). The output of this process provides the data for the training pipeline.

\textbf{Data Exploration}: Data scientists and model developers can search the \textit{Feature Repository} to visualize and extract feature sets  to explore and discover features to be applied to cohort construction. 

\textbf{Feature Generation}: As new features are defined, metadata is added to the catalog and \textit{Feature Generators} that 
    produce the features are associated with the feature metadata.  Feature definition includes specifying dependencies and 
    grouping of features that should be treated as a unit.  

\textbf{Embeddings Generation}: Embeddings may be generated from any subset of available data including from {EHR} 
    which provide demographics data, encounter data, and notes \cite{choi2016multi}. These embedding become available as features and are registered and stored in the \textit{Feature Repository}.
    
\textbf{Training Pipeline}: The training pipeline consists of multiple models that predict the target event, perform model calibration, bias removal, calculate uncertainty metrics, expose feature importance, and perform explainability post-hoc analysis.
    \begin{enumerate}
        \item The modeling process starts with the creation of a model definition in the \textit{Model Registry}.  The model definition contains the 
	complete specification and provenance, including what train and test data were used, model algorithm, hyper-parameters, feature generators, metrics and thresholds. This definition is continuously updated using \textit{Model Registry Client API} as 
	pipeline is executed. 
        \item Multiple \textit{Feature Generators}, as part of the training pipeline, are scaled out and run in parallel. Inter-feature dependencies are taken into consideration in specifying the sequence and parallelization. \textit{Feature Generators} subscribe to feature data in the \textit{Feature Repository} for incremental update.
        \item \textit{Model Registry} tracks the metrics of training experiments run by data scientists and metrics 
        to identify algorithms and parameters that result in the best model performance. At this stage, based on a release management process, the model is promoted to the production environment.
    \end{enumerate}
    
\vspace{1em}\textbf{Integrated AI Life-cycle Management}:
    The architecture facilitates complete or partial automation of key AI life-cycle management activities:
    \begin{enumerate}
        \item \textit{Model Monitoring}: The data from \textit{Feedback Repository} is used by monitoring components for retrospective testing of model accuracy, sensing of model drift and skew. Anomalies and significant changes in accuracy trigger a notification for a data scientist to do further analysis and decide whether to retrain the model.
        \item \textit{AutoML}: AutoML allows AI researchers to automate many of the complicated and time-consuming tasks of feature engineering, model selection and hyper-parameter tuning, to optimize model performance end-to-end.
        \item \textit{Provenance}: Tracking provenance of all aspects of model building is essential for reproducibility. This includes training data, software implementation of the feature generation logic (\textit{Feature Generators}), machine learning algorithm and hyper-parameters used for training. We use DVC library to version the data files which uses the same identifiers as Git\cite{git_2020} source code manager, enabling the unique combined versioning of code and data. Finally, we employ the \textit{Model Registry} to track the model hyperparameters, metrics and models binaries.
    \end{enumerate}
We note that despite the automation, an interdisciplinary team having deep domain knowledge with diverse skills such as machine learning, data analysis and cloud engineering and is required to operate and support the production system.

\section{Results and Conclusions}\label{sec:results}
Currently we have implemented many aspects of the end-to-end system, including the \textit{Model Development Toolkit} and \textit{Model Management} subsystem for the prediction of various health outcomes from medical insurance claims data. We are in the process of operationalizing the \textit{Inference Service} and \textit{Feature Repository}. 

The system was used to analyze a longitudinal medical claim database spanning $\sim220,000$ patient lives covering $\sim330,000$ medical claims. We used a wide variety of features including demographic, diagnosis history, and procedure history of the patient's medical claims to build models for two endpoints. The endpoints were selected with hospital administrators and clinicians as our end users. The scalable nature of the system allowed for semi-concurrent training of more than $10$ architectures, spanning both classical and deep-learning models, over multiple variations/subsets of patient data, covering more than $530$ raw features, culminating in more than $1000$ experiments over a $1$ month period. All of these experiments were traceable from the input data, to the feature generation, and all the way to model training traces and final trained model. The system enabled easy access to various metrics on standardized comparison scenarios to assist in the final promotion of such models for deployment. The system was used collaboratively in a decentralized manner by a team of $>20$ members, including data scientists, ML/AI researchers, and ML/AI engineers, working across various time zones. Our early experiences in using this system has been quite positive and led to a more effective collaboration between different stakeholders that let them focus on their sub-problem while being connected to the overall analysis. 

Our architecture is designed to address the desired characteristics of AI/ML system for Healthcare (see Section~\ref{sec:challenges}) and has empowered us to conduct large scale experiments in a repeatable and reproducible manner. We intend to carry out formal user studies to further investigate the benefits and gaps in our implementation. Our current efforts also include application of the system to many other problems and improve the system over time by pro-actively incorporating feedback from our users.

\appendix

\end{document}